\documentclass{article}

\usepackage{PRIMEarxiv}
\usepackage{iftex}
\ifPDFTeX
    \usepackage[utf8]{inputenc}
\fi
\ifXeTeX
    \AtBeginDocument{}
\fi
\usepackage[T1]{fontenc}
\usepackage[authoryear,round]{natbib}
\setcitestyle{authoryear,round,citesep={;},aysep={,},yysep={;}}
\usepackage{amsmath,amsfonts,bm}

\def\Figref#1{Figure~\ref{#1}}

\def\eqref#1{equation~\ref{#1}}
\def\Eqref#1{Equation~\ref{#1}}
\def\1{\bm{1}}

\DeclareMathAlphabet{\mathsfit}{\encodingdefault}{\sfdefault}{m}{sl}
\SetMathAlphabet{\mathsfit}{bold}{\encodingdefault}{\sfdefault}{bx}{n}

\usepackage{amssymb}
\usepackage{booktabs}
\usepackage[table]{xcolor}
\usepackage{graphicx}
\usepackage{subcaption}
\usepackage{array}
\usepackage{microtype}
\usepackage{url}
\usepackage[
    colorlinks=true,
    citecolor=green!40!black,
    linkcolor=blue!60!black,
    urlcolor=blue!60!black,
    pdfborder={0 0 0}
]{hyperref}

\hypersetup{
    pdftitle={ReWAM: Reciprocal World Action Models for Interactive Autonomous Driving},
    pdfauthor={Benshan Ma, Pei Liu, Ruiguo Zhong, Lang Zhang, Mingyue Feng, Yaonong Wang, Jun Ma}
}

\title{ReWAM: Reciprocal World Action Models for Interactive Autonomous Driving}

\author{
    Benshan Ma\textsuperscript{1},
    Pei Liu\textsuperscript{1},
    Ruiguo Zhong\textsuperscript{1},
    Lang Zhang\textsuperscript{2}\thanks{Project leader.}\\
    \textbf{Mingyue Feng\textsuperscript{2},
    Yaonong Wang\textsuperscript{2},
    Jun Ma\textsuperscript{1,3}}\thanks{Corresponding author.}\\[0.6em]
    \textbf{\textsuperscript{1}The Hong Kong University of Science and Technology (Guangzhou)}\\
    \textbf{\textsuperscript{2}Leapmotor}\\
    \textbf{\textsuperscript{3}The Hong Kong University of Science and Technology}\\[0.3em]
    \textbf{\texttt{bma224@connect.hkust-gz.edu.cn}}\\[0.3em]
    \normalfont Code: \url{https://github.com/LeapWM/rewam}
}

\begin{document}
\maketitle

\begin{abstract}

In interactive scenarios, an autonomous driving system is required to generate ego actions under the influence of other agents' behaviors. Existing World Action Models (WAMs) typically model other agents as components of the world model rather than as decision-makers that fundamentally shape the action of the ego agent, which impairs their performance in dense interaction scenarios.
We introduce Reciprocal World Action Models (\textbf{ReWAM}), a game-theoretic world action modeling framework that captures the reciprocal influence between the ego agent and other agents by representing them as conditional responders whose actions are mutually influenced. We instantiate this framework with a Level-$k$ response hierarchy, where role-specific ego and other action DiTs exchange compact strategy tokens through cross-agent attention while remaining grounded in a shared representation of the future driving world. To learn the response policy of the ego agent from demonstrations, we formulate expert actions as samples from the best response distribution and jointly optimize the entire hierarchy using conditional flow matching. Our framework is evaluated on the NAVSIM dataset and achieves state-of-the-art performance compared to baselines. The improvement is particularly significant in interactive scenarios, validating that modeling reciprocal responses provides a more effective foundation for interaction-aware world action generation.
\end{abstract}

\section{Introduction}
\label{sec:introduction}

Autonomous driving requires an intelligent system to perceive complex traffic scenarios, anticipate their evolution, and plan a safe policy while interacting
with multiple road users whose decisions are mutually dependent.
End-to-end \citep{meanfuser, ipad, goalflow, drivetransformer, drivedpo} autonomous driving simplifies conventional modular pipelines by
jointly learning driving representations and action directly from historical sensor observations.
Recent methods have improved this paradigm through multimodal feature fusion,
goal-conditioned action generation, unified task modeling, and
preference-aligned policy learning.
However, the end-to-end paradigm primarily learns a direct mapping from
observations to action, making it effective at reproducing demonstrated behavior, but action supervision
alone does not ensure the policy acquires an explicit understanding of
the environment and the mechanisms governing its evolution, thereby limiting
its ability to generalize to real-world scenarios that fall outside the
training distribution.

Compared with direct observation-to-action policies, Vision-Language-Action
(VLA) models introduce language as a semantic interpreter between visual
perception and action generation. By aligning visual observations with
linguistic concepts, they are capable of interpreting road semantics, driving rules, and traffic participants' behaviors,  and reasoning about an appropriate driving decision \citep{vla4ad_survey, openvla, omnidrive, opendrivevla, vlme2e, autovla}.
Reasoning-centric extensions further employ structured reasoning, cognitive
feedback, and self-improvement to connect language-based scene understanding
with executable action
\citep{cogdriver, recogdrive, reasoningvla, selfimprovingvla}.
More recent methods combine VLA policies with prospective scene prediction or
latent world modeling, allowing decisions to exploit imagined futures
\citep{worldvla, driveworldvla, vla_world_model, futurevla, flare,
fastinslow}. These advances substantially strengthen semantic scene
understanding and high-level reasoning. However, reasoning in the language
space alone does not explicitly model the continuous physical evolution of the
environment under different actions. A semantically plausible decision may
therefore lack a grounded prediction of its consequences in the future world.

World models address a complementary limitation by learning how driving scenarios
evolve over time. They forecast future observations or latent states
conditioned on scenario history and candidate action, enabling a policy to
evaluate possible consequences before acting
\citep{drivedreamerpolicy, world4drive, reworld, dreamzero}.
World Action Models (WAMs) further couple future generation and action
prediction within a shared architecture, exposing predictive world
representations directly to the planning process
\citep{drivelaw, lawam, latentwam, fastwam, gigaworldpolicy, leworldmodel}.
Many predictive-representation approaches have demonstrated the value of
compact latent dynamics for action-conditioned reasoning
\citep{chainofworld, vlajepa, jepa}. Despite this progress, existing WAMs
primarily model how the ego action influences the future world. Other
agents are usually predicted as part of that world, but their actions are not
explicitly organized as strategic responses to the ego plan. Consequently, an
action-conditioned future can be visually plausible without being behaviorally
consistent with the interaction induced by other agents.

Interaction-aware prediction and planning methods explicitly model this
reciprocal influence through conditional trajectory prediction, joint
planning, or iterative behavioral refinement
\citep{gameformer, ipp, interactiveflow, interaction_sparse_transformer,
interaction_transfer}. However, these methods predominantly operate in
explicit state, maneuver, or action spaces, and their interaction modules
are often separated from generative world modeling. This reveals a gap between
two complementary directions: WAMs learn expressive action-conditioned
representations of future driving worlds but lack an explicit strategic
reasoning structure, whereas interaction-aware prediction and planning model behavioral responses
but do not directly exploit the latent representations learned by generative
world models.

\begin{figure}[!t]
\centering
\captionsetup[subfigure]{skip=0.1em}

\begin{subfigure}[t]{0.495\linewidth}
    \centering
    \begin{minipage}[c][0.12\textheight][c]{\linewidth}
        \centering
        \includegraphics[
            width=\linewidth,
            height=0.12\textheight,
            keepaspectratio
        ]{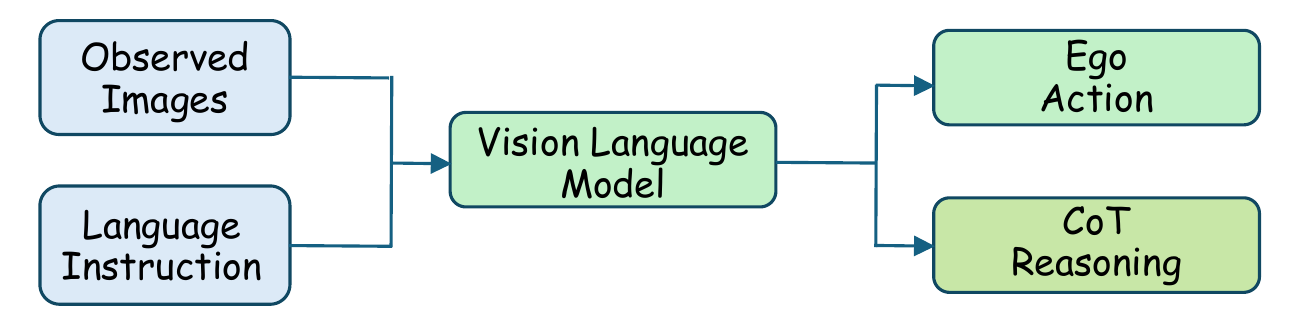}
    \end{minipage}
    \caption{Vision-Language-Action models.}
    \label{fig:paradigm_vla}
\end{subfigure}
\hfill
\begin{subfigure}[t]{0.495\linewidth}
    \centering
    \begin{minipage}[c][0.12\textheight][c]{\linewidth}
        \centering
        \includegraphics[
            width=\linewidth,
            height=0.12\textheight,
            keepaspectratio
        ]{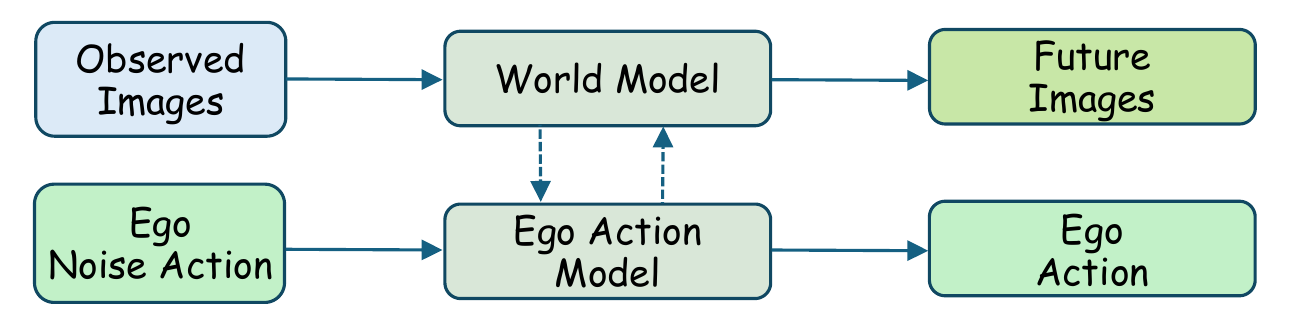}
    \end{minipage}
    \caption{World Action Models.}
    \label{fig:paradigm_wam}
\end{subfigure}

\begin{subfigure}[t]{0.495\linewidth}
    \centering
    \begin{minipage}[c][0.14\textheight][c]{\linewidth}
        \centering
        \includegraphics[
            width=\linewidth,
            height=0.14\textheight,
            keepaspectratio
        ]{\detokenize{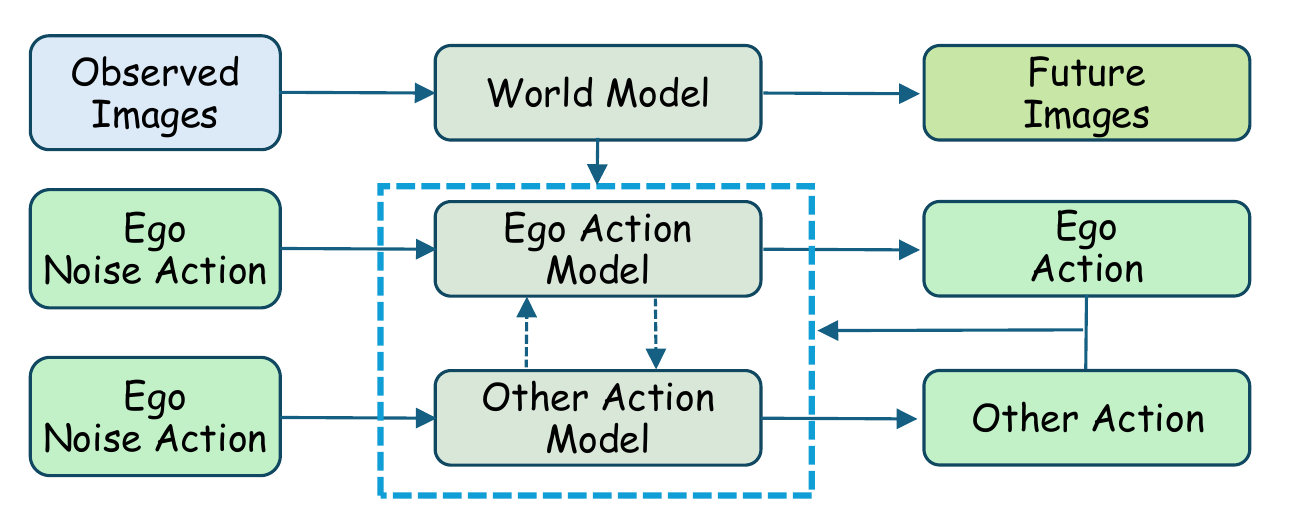}}
    \end{minipage}
    \caption{Reciprocal World Action Model.}
    \label{fig:paradigm_levelk_wam}
\end{subfigure}
\hfill
\begin{subfigure}[t]{0.495\linewidth}
    \centering
    \begin{minipage}[c][0.14\textheight][c]{\linewidth}
        \centering
        \includegraphics[
            width=\linewidth,
            height=0.14\textheight,
            keepaspectratio
        ]{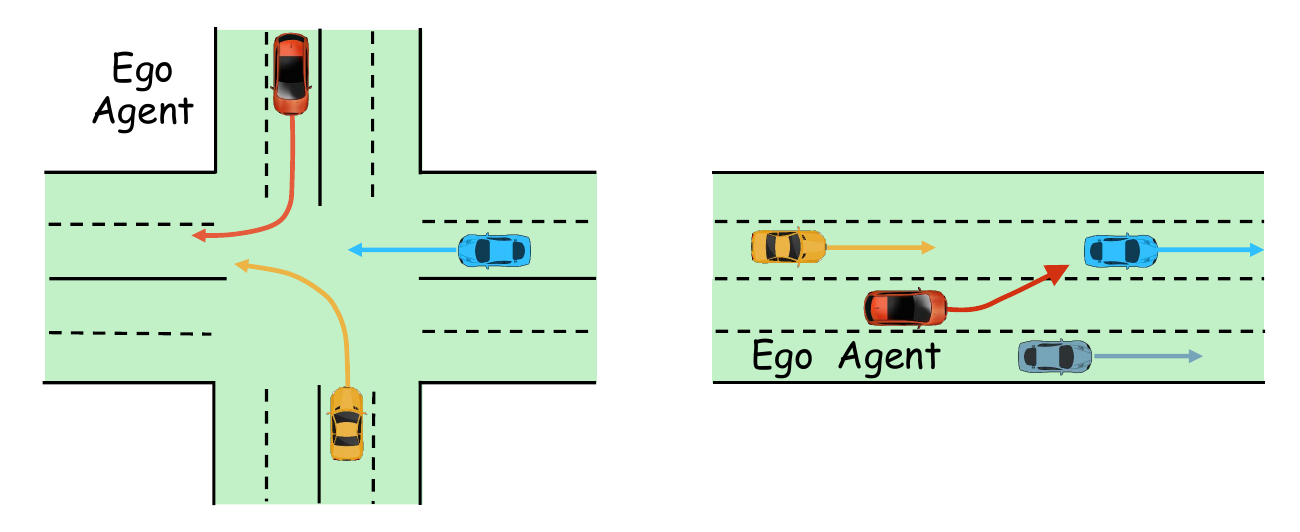}
    \end{minipage}
    \caption{Interactive Driving Scenarios.}
    \label{fig:paradigm_scenes}
\end{subfigure}

\vspace{-0.3em}

\caption{
    Comparison of different autonomous driving paradigms and representative interactive
    driving scenarios. (a) Vision-Language-Action models incorporate
    language-aligned semantic reasoning into action generation.
    (b) World Action Models ground ego-action generation in predicted world
    representations. (c) The proposed Reciprocal World Action Model explicitly captures
    reciprocal behavioral influence through recursive strategic reasoning.
    (d) Representative interactive scenarios in which the behaviors of the ego agent and other agents are mutually influenced.}
\label{fig:paradigm_comparison}
\end{figure}

Bridging these two directions requires more than adding an auxiliary
other-agent prediction head to a WAM. A suitable interaction mechanism
should preserve the learned representation of future world dynamics, represent
ego and other agents' decisions as mutually dependent, and remain
tractable in scenes involving multiple agents. In particular, it should capture
a finite chain of behavioral responses without requiring the computation of a
fully rational multi-agent equilibrium.
A level-$k$ game is a non-equilibrium model of bounded rationality that
organizes agents into a hierarchy of strategic types, making it particularly
well suited to autonomous driving. To date, the level-$k$ game has been applied to lane changing, speed management,
unsignalized intersections, and contingency planning
\citep{karimi_levelk, yuan_levelk, contingency_levelk,
adaptive_rationality}. However, existing formulations generally operate over
handcrafted states, discrete maneuvers, or explicit actions. To the best
of our knowledge, integrating the Level-$k$ strategic hierarchy directly into
the latent representations of a generative WAM remains unexplored. 

In this paper, we propose \textbf{ReWAM}, a game-theoretic framework
that incorporates bounded strategic reasoning into world-action modeling to
explicitly capture how the ego agent and other agents mutually influence one another's actions. Our core idea is to
represent the ego agent and other agents as mutually
responsive decision makers grounded in a shared future-world representation.
At each reasoning level, the ego and other agents' action branches
synchronously update their action hypotheses by conditioning on the joint
strategy predicted at the preceding level. This design retains the expressive
latent dynamics of a WAM while making reciprocal behavioral responses explicit
in action generation.
Our contributions are summarized as follows:

\begin{itemize}
    \item We  explicitly model
    the reciprocal influence between ego and other agents' actions
    within a WAM, extending WAMs beyond one-way ego-action conditioning toward interaction-aware action generation.

    \item We propose ReWAM, a game-theoretic world-action modeling framework that
    integrates bounded reasoning with role-specific action DiTs, enabling the consideration of the reciprocal influence between the ego and other agents' actions.

    \item We evaluate the proposed ReWAM on NAVSIM, where it achieves state-of-the-art (SOTA) performance and delivers particularly
    pronounced improvements in scenarios with dense interactions, which establishes reciprocal response modeling as a critical capability for robust world-action generation in interactive scenarios.
\end{itemize}
\section{Methodology}
\label{sec:method}

In this section, we introduce the proposed  ReWAM framework, a game-theoretic
WAM that augments generative world grounding with finite-depth
strategic reasoning for interactive autonomous driving. Rather than treating
the behavior of other agents as an implicit component of future-world
prediction, ReWAM casts ego and other agents' actions as mutually
conditioned responses within a shared latent representation of the world. As illustrated in
\Figref{fig:proposed_framework}, we formalize this hierarchy as a finite
Level-$k$ game, instantiate its response policies with separate ego and
other-agent action models grounded in a shared world model, and learn the
resulting conditional responses from expert demonstrations.

\subsection{\texorpdfstring{Level-$k$}{Level-k} Game for Solving Interactive Policy}
\label{sec:interactive_problem}

Consider a scenario containing an ego agent $e$ and $N$ other agents,
with participant set $\mathcal{P}=\{e,1,\ldots,N\}$. At time $t$, the model
receives an observation $\mathcal{O}_t$ containing the visual history, textual
navigation condition, and current participant states $s_t$. Let $\tau_i$ denote
the future action of participant $i$, and let
$\tau_{-i}=(\tau_j)_{j\in\mathcal{P}\setminus\{i\}}$ indicate the joint action of all remaining participants. Similarly, $\pi_i$ denotes the policy of
participant $i$, and $\pi_{-i}$ represents the corresponding joint policy of the
remaining participants. Given $\pi_{-i}$, the expected return and best-response set of participant $i$
are defined as
\begin{align}
    J_i(\pi_i,\pi_{-i};\mathcal{O}_t)
    &=
    \mathbb{E}_{\boldsymbol{\tau}\sim(\pi_i,\pi_{-i})}
    \left[
        R_i(\boldsymbol{\tau};\mathcal{O}_t)
    \right],
    \label{eq:expected_return}\\
    \operatorname{BR}_i(\pi_{-i};\mathcal{O}_t)
    &=
    \arg\max_{\pi_i}
    J_i(\pi_i,\pi_{-i};\mathcal{O}_t),
    \label{eq:reward_best_response}
\end{align}
where $\boldsymbol{\tau}=(\tau_i,\tau_{-i})$ is the joint action and
$R_i$ is the reward of participant $i$. Directly coupling all
participants at the same reasoning level produces circular dependencies and
would generally require solving for a joint equilibrium, which is computationally prohibitive for real-time autonomous
driving. Level-$k$ game
instead replaces such simultaneous coupling with a hierarchy of bounded strategic responses:
\begin{equation}
\begin{aligned}
    \pi_i^{(k)}
    &\in
    \operatorname{BR}_i
    \left(
        \pi_{-i}^{(k-1)};\mathcal{O}_t
    \right),\\
    \operatorname{BR}_i
    \left(
        \pi_{-i}^{(k-1)};\mathcal{O}_t
    \right)
    &=
    \arg\max_{\pi_i}
    \mathbb{E}_{
        \boldsymbol{\tau}
        \sim
        (\pi_i,\pi_{-i}^{(k-1)})
    }
    \left[
        R_i(\boldsymbol{\tau};\mathcal{O}_t)
    \right],
    \qquad k\geq1.
\end{aligned}
\label{eq:classical_br}
\end{equation}
where $\pi_i^{(k)}$ denotes participant $i$'s policy at reasoning level $k$. This formulation converts simultaneous mutual dependence into a finite sequence of strategic responses, preserving reciprocal interaction while avoiding the computation of a joint equilibrium.

\subsection{World Representation Conditioned \texorpdfstring{Level-$k$}{Level-k} Reasoning}
\label{sec:level_k_response}

\begin{figure}[tbp]
    \centering
    \includegraphics[
        width=\linewidth,
        height=0.72\textheight,
        keepaspectratio
    ]{\detokenize{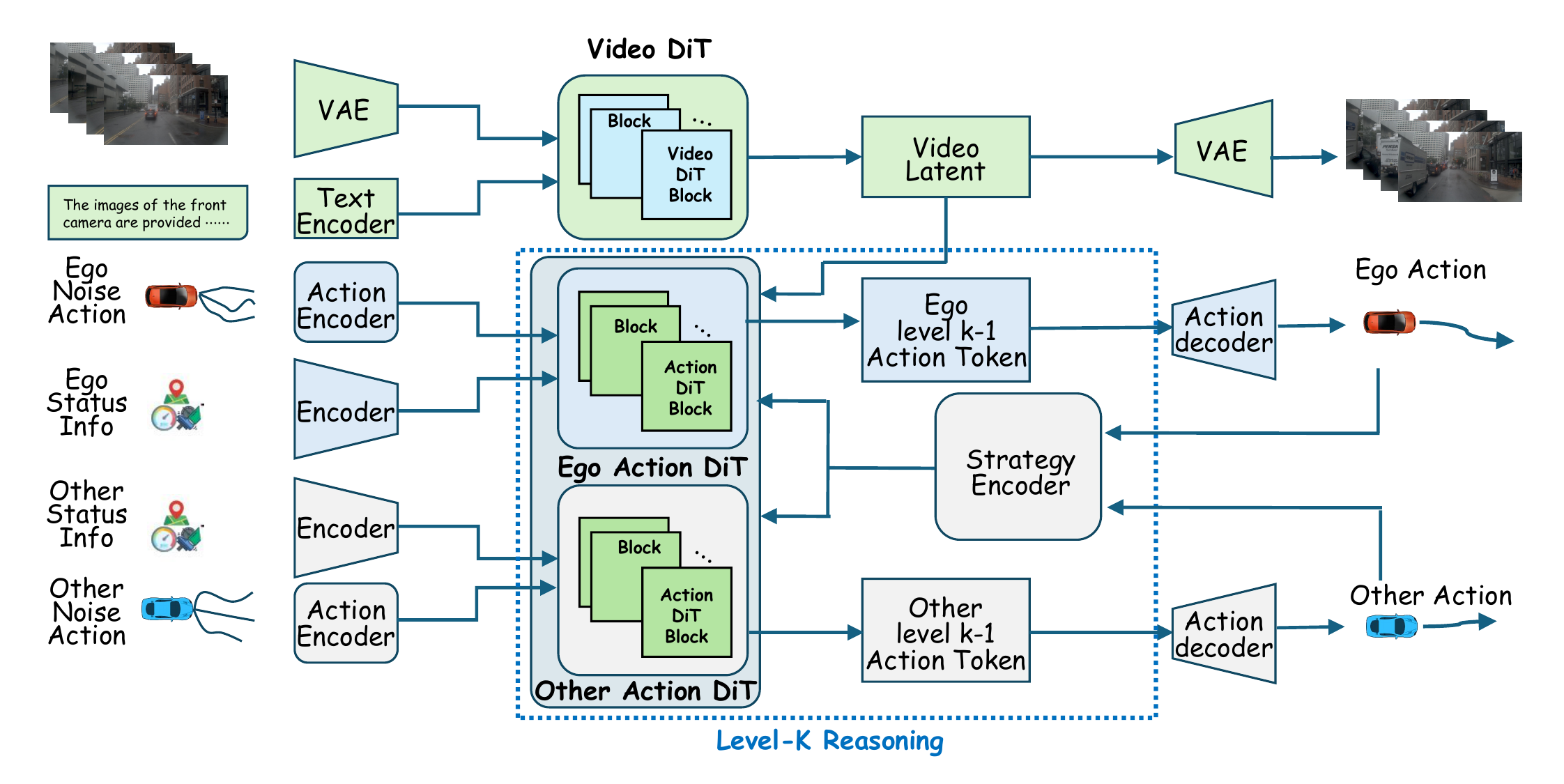}}
    \caption{
        The proposed ReWAM framework.
        Visual observations and language instructions are processed by the
        video branch to construct a shared latent representation of the future
        driving world. Conditioned on this representation and agent-specific
        state information, the Ego and Other Action DiTs synchronously generate
        their action hypotheses. At reasoning level $k$, each branch
        conditions on the level $k-1$ strategies of all remaining valid
        participants, forming a finite hierarchy of reciprocal responses.
    }
    \label{fig:proposed_framework}
\end{figure}

As illustrated in \Figref{fig:proposed_framework}, ReWAM embeds the strategic hierarchy described in ~\eqref{eq:classical_br} into a generative WAM architecture, where a shared world model provides predictive context and dedicated Action DiTs parameterize
the response policies of the ego agent and other agents.
The world model utilizes the visual observations and textual navigation
condition to generate a shared latent representation of the future driving world:
\begin{equation}
    \mathbf{W}_t =
    f_{\phi}^{\mathrm{WM}}(I_{t-H:t},c_t)
    \in\mathbb{R}^{L_v\times d_v},
    \label{eq:world_context}
\end{equation}
where $I_{t-H:t}$ denotes the sequence of historical visual observations up to
time $t$, $c_t$ denotes the textual navigation condition, and $L_v$ and $d_v$
denote the number and dimension of world tokens, respectively. The same world
tokens $\mathbf{W}_t$ are provided to all action branches so that their
responses are grounded in a consistent representation of the driving world.

Grounded in the shared world representation, ReWAM employs two
role-specific response generators: an Ego Action DiT parameterized by
$\theta_e$, which predicts the ego agent's response to the preceding-level
strategies of all other agents, and an Other Action DiT parameterized by
$\theta_o$, which is shared across other agents and predicts each other
agent's response.  Let $\mathbf{Z}_{-i}^{(k-1)}$ denote the strategy
encoder representation of the level $k-1$ action of all participants other than $i$. Conditioned on a realized
preceding-level strategy, the current action hypotheses are sampled as
\begin{align}
    \tau_e^{(k)}
    &\sim
    q_{\theta_e}\!\left(
        \cdot
        \mid
        \mathbf{W}_t,s_t^e,\mathbf{Z}_{-e}^{(k-1)}
    \right),
    \label{eq:ego_recursive_response}\\
    \tau_i^{(k)}
    &\sim
    q_{\theta_o}\!\left(
        \cdot
        \mid
        \mathbf{W}_t,s_t^i,\mathbf{Z}_{-i}^{(k-1)}
    \right),
    \qquad i=1,\ldots,N
    \label{eq:other_recursive_response}
\end{align}
where $(\mathbf{W}_t,s_t^i)$ represents the decision-relevant information in
$\mathcal{O}_t$ for participant $i$. 
Therefore, ~\eqref{eq:ego_recursive_response} and
\eqref{eq:other_recursive_response} provide amortized,
action-conditioned realizations of the bounded responses in
\eqref{eq:classical_br}, where every level $k$ prediction responds to a fixed
realization of the remaining participants' level $k-1$ strategies, rather
than solving a simultaneous equilibrium online.

\subsection{Learning Best Responses from Demonstrations}
\label{sec:response_learning}
Since specifying driving rewards for real-world scenarios is rather challenging, it is difficult to directly compute the best-response policies characterized in 
\eqref{eq:reward_best_response} . We instead assume
that the demonstrated action is generated by a joint expert policy
$\pi^{\mathrm{E}}=(\pi_j^{\mathrm{E}})_{j\in\mathcal{P}}$, whose components
are reward-maximizing responses to the remaining expert policies:
\begin{align}
    \pi_i^{\mathrm{E}}
    &\in
    \operatorname{BR}_i
    \left(
        \pi_{-i}^{\mathrm{E}};\mathcal{O}_t
    \right),
    \label{eq:expert_best_response}\\
    \boldsymbol{\tau}^{\mathrm{demo}}
    &\sim
    \pi^{\mathrm{E}}\!\left(
        \cdot\mid\mathcal{O}_t
    \right),
    \label{eq:expert_demonstration}
\end{align}
where $\boldsymbol{\tau}^{\mathrm{demo}}$ is the demonstrated joint action.
Let $p_{\mathrm{demo}}(\tau_i\mid\mathcal{O}_t,\tau_{-i})$ denote the
conditional response distribution induced by the joint demonstrations. After
applying the world, state, and strategy encoders, we use the same notation for
the corresponding model-level conditional distribution
$p_{\mathrm{demo}}(\tau_i\mid\mathbf{W}_t,s_t^i,\mathbf{Z}_{-i})$.
The response model can be learned through conditional distribution matching:
\begin{equation}
    \theta_i^{\star}
    =
    \arg\min_{\theta_i}
    \mathcal{D}\!\left(
        p_{\mathrm{demo}}\!\left(
            \tau_i
            \mid
            \mathbf{W}_t,s_t^i,\mathbf{Z}_{-i}
        \right),
        q_{\theta_i}\!\left(
            \tau_i
            \mid
            \mathbf{W}_t,s_t^i,\mathbf{Z}_{-i}
        \right)
    \right),
    \label{eq:response_distribution_matching}
\end{equation}
where $\theta_i=\theta_e$ for the ego vehicle and $\theta_i=\theta_o$ for other agents, respectively;  and $\mathcal{D}$ measures the discrepancy between the
demonstrated and learned conditional distributions. The resulting
$q_{\theta_i}$ is therefore an amortized, stochastic surrogate for the
best-response policy over the support of the demonstrated strategy
distribution.

\subsection{\texorpdfstring{Level-$k$}{Level-k} Interaction Block}
\label{sec:strategy_conditioned_generation}

\begin{figure}[t]
    \centering
    \includegraphics[
        width=0.8\linewidth,
        height=0.60\textheight,
        keepaspectratio
    ]{\detokenize{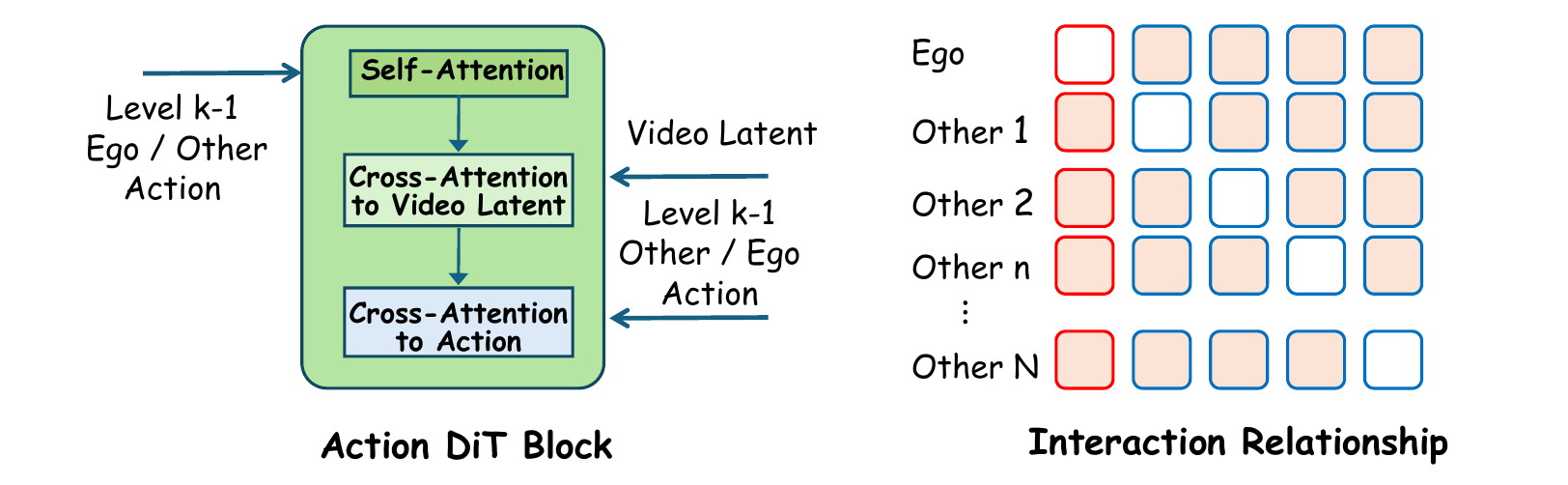}}
    \caption{
        Architecture of the proposed Level-$k$ interaction block.
        Within each Action DiT block, temporal self-attention models the target agent's action sequence, while cross-attention to the Video DiT features grounds action generation in the predicted driving world.
        Cross-agent attention further conditions the current response on the preceding-level action hypotheses of the remaining agents. The elaborated Action DiT architecture explicitly captures reciprocal strategic influence among agents in interactive scenarios.
    }
    \label{fig:interaction_block}
\end{figure}

As illustrated in \Figref{fig:proposed_framework}, ReWAM parameterizes
the response distributions of the ego agent and other agents using two
Action DiTs that progressively transform noisy action sequences into
world-grounded and interaction-conditioned action predictions through stacked
transformer blocks. Each block
successively applies temporal self-attention over the target participant's
future action tokens, and cross-attention to the corresponding Video DiT features,
cross-agent attention to the preceding-level action hypotheses, and a
feed-forward transformation. This architecture integrates temporal action modeling, world-grounded contextualization, and explicit strategic
interaction within one WAM framework.
As shown in \Figref{fig:interaction_block}, to model the interaction among agents, the cross-attention for participant $i$ at transformer block $\ell$ is
computed as
\begin{equation}
    \mathbf{A}_{i,\ell}^{(k)}
    =
    \operatorname{softmax}\!\left(
        \frac{
            Q_{\ell}(\mathbf{H}_{i,\ell}^{(k)})
            K_{\ell}(\mathbf{Z}^{(k-1)})^{\top}
        }{\sqrt{d_h}}
        +
        \mathbf{M}_i
    \right)
    V_{\ell}(\mathbf{Z}^{(k-1)}),
    \label{eq:cross_agent_attention}
\end{equation}
where $\mathbf{H}_{i,\ell}^{(k)}\in\mathbb{R}^{T\times d}$ denotes the action
tokens of participant $i$; $d_h$ is the attention-head dimension; the target participant's action tokens serve as queries $Q_{\ell}$, while the
strategy representations of the preceding-level action provide keys $K_{\ell}$ and values $V_{\ell}$.
The mask $\mathbf{M}_i$ imposes role-dependent visibility by suppressing
padded participants and the target participant's own strategy token.
Therefore, the proposed interaction blocks enable cross-agent attention to explicitly model the mutual influence among participants by conditioning each participant's
current response on the preceding-level action hypotheses of all other valid
participants.

\subsection{Joint Learning and Inference}
\label{sec:joint_learning}
Flow matching provides a tractable velocity-regression
formulation of the distribution matching objective,  so it is adopted to learn the response model targeted by \eqref{eq:response_distribution_matching}. For participant
$j\in\mathcal{P}$ at reasoning level $k$, let
$\mathbf{x}_{0,j}$ denote a normalized demonstrated action sampled from the
target conditional distribution:
\begin{equation}
\mathbf{x}_{0,j}
\sim
p_{\mathrm{demo}}\!\left(
    \cdot
    \mid
    \mathbf{W}_t,
    s_t^j,
    \mathbf{Z}_{-j}^{(k-1)}
\right).
\label{eq:conditional_demonstration}
\end{equation}
Then, we sample $\boldsymbol{\epsilon}\sim\mathcal{N}(\mathbf{0},\mathbf{I})$ and
$r\sim\mathcal{U}(0,1)$, and define
\begin{align}
\mathbf{x}_{r,j}
&=
(1-r)\mathbf{x}_{0,j}
+
r\boldsymbol{\epsilon},
\label{eq:compact_flow_path}\\
\mathbf{u}_{r,j}
&=
\boldsymbol{\epsilon}-\mathbf{x}_{0,j}.
\label{eq:compact_flow_velocity}
\end{align}
where $\boldsymbol{\epsilon}$ indicates a Gaussian noise sample that defines the source distribution; $r$ is the continuous flow time that interpolates
between the demonstrated action and noise; $\mathbf{x}_{r,j}$ denotes the action state at flow time $r$ along the
linear interpolation between the demonstrated action and Gaussian noise;
$\mathbf{u}_{r,j}$ represents the corresponding target velocity.
The resulting state-velocity pairs provide the regression targets for the conditional velocity field, which is learned by minimizing
\begin{equation}
    \mathcal{L}_{j}^{(k)}
    =
    \mathbb{E}\!\left[
        \left\|
        v_{\theta_j}\!\left(
            \mathbf{x}_{r,j},r
            \mid
            \mathbf{W}_t,s_t^j,\mathbf{Z}_{-j}^{(k-1)}
        \right)
        -
        \mathbf{u}_{r,j}
        \right\|_2^2
    \right].
    \label{eq:participant_flow_matching}
\end{equation}

\noindent
\Eqref{eq:participant_flow_matching} defines the participant-level
flow-matching objective for a single response at reasoning level $k$, and these
objectives are subsequently aggregated according to the two response roles:
\begin{equation}
\mathcal{L}_{\mathrm{ego}}^{(k)}
=
\left.\mathcal{L}_j^{(k)}\right|_{j=e},
\qquad
\mathcal{L}_{\mathrm{other}}^{(k)}
=
\frac{
    \sum_{i=1}^{N}m_i\mathcal{L}_i^{(k)}
}{
    \sum_{i=1}^{N}m_i
},
\label{eq:branch_flow_losses}
\end{equation}
where $m_i\in\{0,1\}$ indicates the validity of surrounding agent $i$.
The branch-level objectives are aggregated across the reasoning hierarchy
using normalized depth-dependent weights:
\begin{equation}
\mathcal{L}_{\mathrm{flow}}
=
\sum_{k=1}^{K_{\mathrm{train}}}
\alpha_k
\left(
\lambda_{\mathrm{ego}}\mathcal{L}_{\mathrm{ego}}^{(k)}
+
\lambda_{\mathrm{other}}\mathcal{L}_{\mathrm{other}}^{(k)}
\right),
\qquad
\alpha_k
=
\frac{k^p}{\sum_{\ell=1}^{K_{\mathrm{train}}}\ell^p}.
\label{eq:compact_total_loss}
\end{equation}
where $K_{\mathrm{train}}$ specifies the supervised reasoning depth; $\alpha_k$ denotes the normalized depth-dependent weight;
$p$ controls the relative emphasis assigned to deeper response levels; $\lambda_{\mathrm{ego}}$ and $\lambda_{\mathrm{other}}$ balance the two action
branches; the normalization ensures
$\sum_{k=1}^{K_{\mathrm{train}}}\alpha_k=1$.
Accordingly, the response models are jointly optimized under the aggregate
objective:
\begin{equation}
\Theta^{\star}
\in
\arg\min_{\Theta}
\mathcal{L}_{\mathrm{flow}}(\Theta),
\label{eq:flow_matching_optimum}
\end{equation}
where $\Theta$ denotes the complete set of trainable response-model parameters. At the population optimum, conditional flow matching recovers the
demonstration-conditioned response distribution for each valid participant and
supervised reasoning level, which means
\begin{equation}
q_{\theta_j^{\star}}\!\left(
    \cdot
    \mid
    \mathbf{W}_t,s_t^j,\mathbf{Z}_{-j}^{(k-1)}
\right)
=
p_{\mathrm{demo}}\!\left(
    \cdot
    \mid
    \mathbf{W}_t,s_t^j,\mathbf{Z}_{-j}^{(k-1)}
\right),
\qquad
j\in\mathcal{P},\quad
k=1,\ldots,K_{\mathrm{train}}.
\label{eq:flow_matching_consistency}
\end{equation}
where $\theta_j^{\star}=\theta_e^{\star}$ for the ego agent and
$\theta_j^{\star}=\theta_o^{\star}$ for other agents. Therefore, the distribution-matching objective in
\eqref{eq:response_distribution_matching} is achieved  across response roles and
reasoning levels by conditional flow matching.

\begin{table}[tbp]
    \centering
    \caption{Comparison with state-of-the-art methods on the NAVSIM benchmark.
    Higher values indicate better performance.}
    \label{tab:navsim_comparison}
    \small
    \setlength{\tabcolsep}{3pt}
    \renewcommand{\arraystretch}{1.02}
    \begin{tabular*}{\linewidth}{@{\extracolsep{\fill}}l|c|ccccc|c@{}}
        \toprule
        \textbf{Method}
        & \textbf{Ref}
        & $\mathbf{NC}\!\uparrow$
        & $\mathbf{DAC}\!\uparrow$
        & $\mathbf{TTC}\!\uparrow$
        & $\mathbf{Comf.}\!\uparrow$
        & $\mathbf{EP}\!\uparrow$
        & $\mathbf{PDMS}\!\uparrow$ \\
        \midrule

        \multicolumn{8}{@{}l}{\textit{Traditional End-to-End Methods}} \\
        VADv2-$V_{8192}$ ~\citep{VADv2}
        & arXiv'24
        & 97.2 & 89.1 & 91.6 & \textbf{100} & 76.0
        & 80.9 \\

        UniAD ~\citep{UniAD}
        & CVPR'23
        & 97.8 & 91.9 & 92.9 & \textbf{100} & 78.8
        & 83.4 \\

        TransFuser ~\citep{TransFuser}
        & TPAMI'23
        & 97.7 & 92.8 & 92.8 & \textbf{100} & 79.2
        & 84.0 \\

        PARA-Drive ~\citep{PARA-Drive}
        & CVPR'24
        & 97.9 & 92.4 & 93.0 & 99.8 & 79.3
        & 84.0 \\

        DiffusionDrive ~\citep{DiffusionDrive}
        & CVPR'25
        & 98.2 & 96.2 & 94.7 & \textbf{100} & 82.2
        & 88.1 \\

        ReCogDrive-IL ~\citep{recogdrive}
        & ICLR'26
        & 98.1 & 94.7 & 94.2 & \textbf{100} & 80.9
        & 86.5 \\

        \midrule
        \multicolumn{8}{@{}l}{\textit{World Model Methods}} \\

        DrivingGPT ~\citep{DrivingGPT}
        & ICCV'25
        & 98.9 & 90.7 & 94.9 & 95.6 & 79.7
        & 82.4 \\

        LAW ~\citep{LAW}
        & ICLR'25
        & 96.4 & 95.4 & 88.7 & 99.9 & 81.7
        & 84.6 \\

        Epona ~\citep{Epona}
        & ICCV'25
        & 97.9 & 95.1 & 93.8 & 99.9 & 80.4
        & 86.2 \\

        WoTE ~\citep{WoTE}
        & ICCV'25
        & 98.5 & 96.8 & 94.9 & 99.9 & 81.9
        & 88.3 \\

        DriveVLA-W0 ~\citep{drivevla}
        & ICLR'26
        & 98.4 & 95.3 & 95.2 & \textbf{100} & 80.9
        & 87.2 \\

        PWM ~\citep{PWM}
        & NeurIPS'25
        & 98.6 & 95.9 & 95.4 & \textbf{100} & 81.8
        & 88.1 \\

        WorldDrive ~\citep{WorldDrive}
        & arXiv'26
        & 98.4 & 96.8 & 95.2 & \textbf{100} & 83.3
        & 89.0 \\

        DriveLaW ~\citep{drivelaw}
        & CVPR'26
        & 99.0 & 97.1 & 96.7 & \textbf{100} & 81.3
        & 89.1 \\

        ReWorld ~\citep{reworld}
        & arXiv'26
        & 99.1 & \textbf{98.2} & \textbf{97.7} & \textbf{100} & 82.0
        & 90.4 \\

        ReWAM (Ours)
        & -
        & \textbf{99.3}
        & 97.8
        & 97.2
        & \textbf{100}
        & \textbf{83.6}
        & \textbf{90.6} \\
        \bottomrule
    \end{tabular*}
\end{table}

\section{Experiments}
\label{sec:experiments}

\subsection{Experimental Setup}
\label{sec:experimental_setup}

\paragraph{Implementation details.}
We implement the proposed ReWAM on top of DriveLaW~\citep{drivelaw} by designing the Level-$k$ interaction block and introducing a parameter-shared Other Action DiT together with a strategy encoder. The pretrained VAE, text encoder, and Video DiT are
frozen, while both Action DiTs and the associated interaction modules are
optimized. We first train the Level-1 model and initialize every
$k\geq2$ model from the resulting checkpoint for further fine-tuning. We use
four front-camera frames sampled at 2\, Hz and predict eight poses at 0.5\,s
intervals for the ego agent and at most ten other agents within 80\,m.
Current agent states are obtained from scene annotations, whereas
future agent states and actions are used only as supervision.

\paragraph{Dataset and evaluation metrics.}
We conduct experiments on NAVSIM, a planning-oriented benchmark derived from
the public OpenScene/nuPlan driving logs. We use the official \texttt{navtrain}
split for training and the \texttt{navtest} split for evaluation. The latter
contains 136 logs and 12,146 driving scenarios. Unless stated otherwise, we
report the Predictive Driver Model Score (PDMS) together with its components:
No-at-Fault Collision (NC), Drivable Area Compliance (DAC), Time to Collision
(TTC), Comfort, and Ego Progress (EP). We additionally report Driving
Direction Compliance (DDC) as a diagnostic measure. All metric values are
multiplied by 100, and higher values indicate better performance.

\subsection{Main Results}
\label{sec:main_results}

\paragraph{Comparison with SOTA methods.}

As presented in Table~\ref{tab:navsim_comparison}, ReWAM achieves the highest PDMS score of 90.6, improving over its
DriveLaW backbone by 1.5 points and the SOTA method, ReWorld,
by 0.2 points. The gain over DriveLaW is accompanied by improvements in NC,
DAC, TTC, and EP scores, with the largest enhancement appearing in EP score. Although
ReWorld obtains slightly higher DAC and TTC scores, ReWAM achieves better performance in NC and EP scores and consequently a better aggregate score. These results validate that explicitly modeling the interaction between the
ego agent and other agents enables the planner to better anticipate reciprocal
behavioral responses, thereby achieving a SOTA performance on the benchmark.

\subsection{Analysis of \texorpdfstring{Level-$k$}{Level-k} Reasoning}
\label{sec:level_k_analysis}

\paragraph{Effect of reasoning depth}
We vary the reasoning depth while keeping the model and evaluation protocol
fixed, and report the performance at each reasoning level in
Table~\ref{tab:interaction_level_comparison}. Increasing the depth from
Level-1 to Level-4 consistently improves the aggregate performance, raising
PDMS from 90.03 to 90.64. The largest single-level gain occurs at Level-2,
whereas subsequent levels deliver progressively smaller improvements. This
trend is most pronounced in TTC, DAC, and EP, indicating that recursive
response reasoning enhances collision avoidance, drivable-area compliance,
and driving progress, while Comfort and DDC remain consistently high across
all depths. Although Level-5 marginally
improves NC and EP, it yields no further gain in PDMS and slightly reduces DAC.
These results suggest that deeper reasoning progressively captures strategic
dependencies among traffic participants, but exhibits diminishing returns once
the dominant interaction structure has been resolved.

\paragraph{Benefits across interaction intensities.}
\label{sec:interaction_intensity}

To assess whether Level-$k$ reasoning yields greater benefits in interactive
settings, we partition all valid tokens into five
interaction-intensity groups (None, Weak, Medium, Strong, and Critical) using a model-independent score derived from the
official reference trajectories. As shown in Table~\ref{tab:interaction_intensity_performance}, ReWAM consistently outperforms both baselines across all five groups,
with its advantage becoming particularly pronounced in the most
interaction-intensive scenarios. Specifically, its improvement over DriveLaW
increases from $1.52$ PDMS scores in the None group to $4.59$  in the
Critical group, where it also surpasses DiffusionDrive by $2.54$. Such
concentration of performance gains in challenging interactive scenarios
provides strong empirical support for the effectiveness of explicitly modeling
reciprocal responses among traffic participants.

\paragraph{Performance decomposition under dense interactions.}
We further examine the Critical
subset, and the results are shown in Table~\ref{tab:critical_interaction_performance}. ReWAM exhibits component-wise dominance over both baselines across all non-saturated metrics,
while retaining the same perfect Comfort score. The concurrent improvements in
NC and TTC indicate a stronger capacity to anticipate and resolve
spatiotemporal motion conflicts. Importantly, these safety gains are accompanied
by higher DAC, DDC, and EP, suggesting that conflict avoidance is achieved
without resorting to off-road, directionally inconsistent, or excessively
conservative behavior. The resulting PDMS improvement therefore reflects a
more favorable coordination of safety, traffic-rule compliance, and driving
progress, rather than optimization of any single metric. This consistent
performance profile provides further evidence that explicit reciprocal-response
reasoning is particularly effective when participants' decisions are strongly
interdependent.

\begin{table}[!t]
    \centering
    \caption{The performance of the proposed method across different reasoning levels.}
    \label{tab:interaction_level_comparison}

    \small
    \setlength{\tabcolsep}{3pt}
    \renewcommand{\arraystretch}{1.02}

    \begin{tabular*}{\linewidth}{
        @{\extracolsep{\fill}}
        l|cccccc|c}
        \toprule
        \textbf{Interaction Level}
        & \textbf{NC}
        & \textbf{DAC}
        & \textbf{TTC}
        & \textbf{Comfort}
        & \textbf{EP}
        & \textbf{DDC}
        & \textbf{PDMS} \\
        \midrule

        Level-1
        & 99.06
        & 97.39
        & 96.25
        & 99.97
        & 83.28
        & 98.12
        & 90.03 \\

        Level-2
        & 99.25
        & 97.68
        & 97.13
        & 99.97
        & 83.44
        & \textbf{98.17}
        & 90.48 \\

        Level-3
        & 99.26
        & 97.74
        & 97.13
        & \textbf{99.98}
        & 83.54
        & 98.15
        & 90.55 \\

        Level-4
        & 99.25
        & \textbf{97.81}
        & \textbf{97.18}
        & 99.96
        & 83.60
        & 98.14
        & \textbf{90.64} \\

        Level-5
        & \textbf{99.29}
        & 97.77
        & \textbf{97.18}
        & 99.97
        & \textbf{83.64}
        & 98.16
        & \textbf{90.64} \\

        \bottomrule
    \end{tabular*}
\end{table}

\begin{table}[!t]
    \centering
    \caption{
        PDMS scores across interaction-intensity groups on NAVTEST.
        Share denotes the percentage of the $12{,}144$ evaluated tokens, and parentheses report ReWAM's absolute improvement over each baseline.}
    \label{tab:interaction_intensity_performance}

    \small
    \setlength{\tabcolsep}{3pt}
    \renewcommand{\arraystretch}{1.02}

    \begin{tabular*}{\linewidth}{
        @{\extracolsep{\fill}}
        lrrrrr
        @{}
    }
        \toprule
        \textbf{Interaction Intensity}
        & \textbf{Tokens}
        & \textbf{Share (\%)}
        & \textbf{ReWAM}
        & \textbf{DriveLaW}
        & \textbf{DiffusionDrive} \\
        \midrule

        None
        & 6621
        & 54.52
        & \textbf{89.92}
        & 88.40 $(+1.52)$
        & 87.06 $(+2.86)$ \\

        Weak
        & 1841
        & 15.16
        & \textbf{90.89}
        & 89.76 $(+1.13)$
        & 89.20 $(+1.69)$ \\

        Medium
        & 1841
        & 15.16
        & \textbf{92.33}
        & 91.73 $(+0.60)$
        & 89.56 $(+2.77)$ \\

        Strong
        & 1288
        & 10.61
        & \textbf{92.43}
        & 90.50 $(+1.93)$
        & 90.17 $(+2.26)$ \\

        Critical
        & 553
        & 4.55
        & \textbf{88.72}
        & 84.13 $(+4.59)$
        & 86.18 $(+2.54)$ \\

        \bottomrule
    \end{tabular*}
\end{table}

\begin{table}[!t]
    \centering
    \caption{
        Performance of the proposed method and baselines on
        critical-interaction scenarios.
    }
    \label{tab:critical_interaction_performance}

    \small
    \setlength{\tabcolsep}{3pt}
    \renewcommand{\arraystretch}{1.02}

    \begin{tabular*}{\linewidth}{
    @{\extracolsep{\fill}}
    l|cccccc|c
    @{\hspace{\tabcolsep}}
}
        \toprule
        \textbf{Method}
        & \textbf{NC}
        & \textbf{DAC}
        & \textbf{TTC}
        & \textbf{Comfort}
        & \textbf{EP}
        & \textbf{DDC}
        & \textbf{PDMS} \\
        \midrule

        ReWAM
        & \textbf{96.93}
        & \textbf{99.10}
        & \textbf{85.90}
        & \textbf{100}
        & \textbf{89.59}
        & \textbf{98.19}
        & \textbf{88.72} \\

        DriveLaW
        & 94.21
        & 98.55
        & 79.20
        & \textbf{100}
        & 87.06
        & 98.01
        & 84.13 \\

        DiffusionDrive
        & 95.39
        & 97.47
        & 84.27
        & \textbf{100}
        & 87.07
        & 97.38
        & 86.18 \\

        \bottomrule
    \end{tabular*}
\end{table}

\begin{figure}[!t]
    \centering

    \begin{subfigure}[t]{0.32\linewidth}
        \centering
        \includegraphics[
            width=\linewidth,
            keepaspectratio
        ]{\detokenize{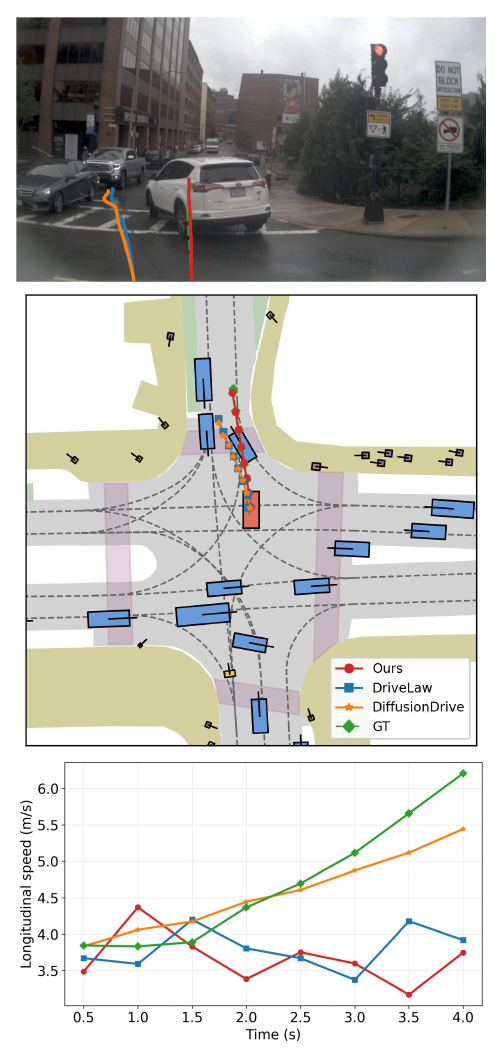}}
        \caption{Narrow Road}
        \label{fig:qualitative_narrow_road}
    \end{subfigure}
    \hfill
    \begin{subfigure}[t]{0.32\linewidth}
        \centering
        \includegraphics[
            width=\linewidth,
            keepaspectratio
        ]{\detokenize{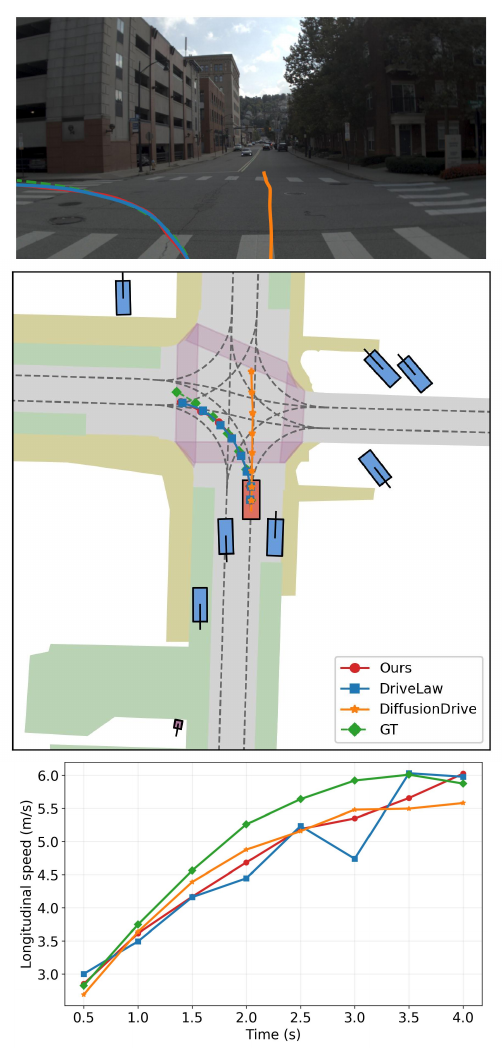}}
        \caption{Left Turn}
        \label{fig:qualitative_left_turn}
    \end{subfigure}
    \hfill
    \begin{subfigure}[t]{0.32\linewidth}
        \centering
        \includegraphics[
            width=\linewidth,
            keepaspectratio
        ]{\detokenize{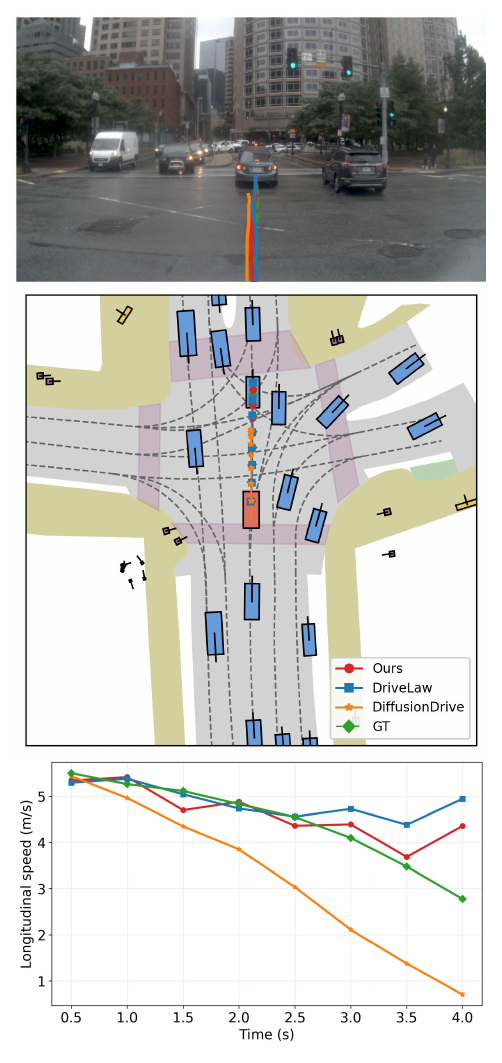}}
        \caption{Going Straight}
        \label{fig:qualitative_going_straight}
    \end{subfigure}

    \caption{
Qualitative comparison across three representative interactive driving
scenarios. Each panel presents the front-camera action projection,
bird's-eye-view trajectory visualization, and longitudinal speed profile.
}
    \label{fig:qualitative_scenarios}
\end{figure}

\subsection{Qualitative Results}
\label{sec:qualitative_results}
 The results of three interactive scenarios, including front-camera-view action projection, bird's-eye-view action visualization, and longitudinal speed profile,  are shown in  \Figref{fig:qualitative_scenarios}. 
 In the narrow road scenario, ReWAM successfully negotiates the constrained passage,
whereas both baselines fail to complete the traversal. 
In the left turn scenario, DiffusionDrive maintains an almost straight trajectory and misses
the intended maneuver, while ReWAM conforms to the turning geometry and preserves smooth progress. 
The going straight scenario further exposes two opposing failure modes: DiffusionDrive decelerates excessively and sacrifices ego progress, whereas DriveLaW responds insufficiently and collides
with the leading vehicle. By contrast, ReWAM adopts a properly calibrated trajectory and speed profile that avoids the conflict without
unnecessary loss of progress. These examples collectively demonstrate the superior performance of
ReWAM in interactive driving scenarios, where reciprocal response
modeling yields safer and more effective action.

\section{Conclusion}
\label{sec:conclusion}

In this paper, we presented the proposed ReWAM, a game-theoretic world-action modeling framework that explicitly captures the reciprocal influence
between the action of the ego agent and other agents, advancing WAMs
from one-way ego-action conditioning toward strategically coupled multi-agent
action generation. The proposed formulation replaces simultaneous multi-agent
coupling with a finite hierarchy of bounded responses, thereby introducing
structured strategic reasoning without incurring online equilibrium
computation. Within this hierarchy, the Ego and parameter-shared Other Action
DiTs are grounded in a shared future-world representation and exchange
preceding-level action hypotheses through cross-agent attention. Conditional
flow matching then learns the corresponding response distributions
directly from demonstrations. ReWAM achieves the best
performance on NAVSIM, with a PDMS of 90.6, while the study on reasoning depth 
shows that recursive updates provide consistent gains before saturating at
deeper levels. Interaction-stratified evaluation further demonstrates that our
method outperforms both DriveLaW and DiffusionDrive across all intensity groups,
with its largest advantage over the DriveLaW backbone arising in the critical
interaction group. The accompanying improvements in safety, traffic-rule
compliance, and ego progress, together with the qualitative examples, validate
that these gains arise from more effective interaction-aware action generation
rather than conservative planning alone. These findings establish bounded
reciprocal-response reasoning as a practical and effective foundation for
interactive world-action modeling.

% Read the verified bibliography directly for portable Overleaf/arXiv builds.
% The editable BibTeX database is retained in references.bib.
% Formatted bibliography snapshot generated from references.bib.
% Regenerate this file after changing bibliography entries or citation keys.

\end{document}